\documentclass[conference]{IEEEtran}
\IEEEoverridecommandlockouts
\usepackage{cite}
\usepackage{amsmath,amssymb,amsfonts}
\usepackage{algorithmic}
\usepackage{graphicx}
\usepackage{textcomp}
\usepackage{xcolor}
\usepackage{fancyhdr}

\usepackage{tablefootnote}
\usepackage{booktabs}
\usepackage{colortbl}
\usepackage{arydshln}
\def\BibTeX{{\rm B\kern-.05em{\sc i\kern-.025em b}\kern-.08em
    T\kern-.1667em\lower.7ex\hbox{E}\kern-.125emX}}

\begin{document}

\pagestyle{fancy}
\chead{\fontsize{7}{12} \textbf{This paper has been accepted at 2023 International Joint Conference on Neural Networks (IJCNN). This is the accepted version. \\ Please find the published version and info to cite the paper at https://doi.org/10.1109/IJCNN54540.2023.10191452}}

\title{To Whom are You Talking? A Deep Learning Model to Endow Social Robots with Addressee Estimation Skills\\
\thanks{For research leading to these results Carlo Mazzola has received funding from the project titled TERAIS in the framework of the program Horizon-Widera-2021 of the European Union under the Grant agreement number 101079338; Marta Romeo from the UKRI Node on Trust (EP/V026682/1); Angelo Cangelosi from the EU projects TRAINCREASE and MUSAE, and the US project THRIVE++.}
}

\makeatletter
\newcommand{\linebreakand}{%
  \end{@IEEEauthorhalign}
  \hfill\mbox{}\par
  \mbox{}\hfill\begin{@IEEEauthorhalign}
}
\makeatother

\author{
\IEEEauthorblockN{1\textsuperscript{st} Carlo Mazzola}
\IEEEauthorblockA{\textit{CONTACT Unit} \\ \textit{Istituto Italiano di Tecnologia} \\ 
\textit{DIBRIS, University of Genoa}\\
Genoa, Italy \\
ORCID 0000-0002-9282-9873}
\and

\IEEEauthorblockN{2\textsuperscript{nd} Marta Romeo}
\IEEEauthorblockA{\textit{School of Mathematical} \\
\textit{and Computer Sciences}\\
\textit{Heriot-Watt University}\\
Edinburgh, United Kingdom \\
ORCID 0000-0003-4438-0255}
\and

\IEEEauthorblockN{3\textsuperscript{rd} Francesco Rea}
\IEEEauthorblockA{\textit{CONTACT Unit} \\
\textit{Istituto Italiano di Tecnologia}\\
Genoa, Italy \\
ORCID 0000-0001-8535-223X}
\linebreakand

\IEEEauthorblockN{4\textsuperscript{th} Alessandra Sciutti}
\IEEEauthorblockA{\textit{CONTACT Unit} \\
\textit{Istituto Italiano di Tecnologia}\\
Genoa, Italy \\
ORCID 0000-0002-1056-3398}
\and

\IEEEauthorblockN{5\textsuperscript{th} Angelo Cangelosi}
\IEEEauthorblockA{\textit{Manchester Centre for Robotics and AI} \\
\textit{University of Manchester}\\
Manchester, United Kingdom \\
ORCID 0000-0002-4709-2243}
}

\maketitle
\thispagestyle{fancy}
\begin{abstract}
Communicating shapes our social word. For a robot to be considered social and being consequently integrated in our social environment it is fundamental to understand some of the dynamics that rule human-human communication. In this work, we tackle the problem of Addressee Estimation, the ability to understand an utterance's addressee, by interpreting and exploiting non-verbal bodily cues from the speaker. We do so by implementing an hybrid deep learning model composed of convolutional layers and LSTM cells taking as input images portraying the face of the speaker and 2D vectors of the speaker's body posture. Our implementation choices were guided by the aim to develop a model that could be deployed on social robots and be efficient in ecological scenarios. We demonstrate that our model is able to solve the Addressee Estimation problem in terms of addressee localisation in space, from a robot ego-centric point of view.
\end{abstract}


\section{Introduction}
Communicating means sharing, might it be a message, a thought, or an inner state, and is an act that inherently shapes the social world. 
To properly be part of the social environment, each agent needs to understand some basic dynamics of communication, such as to whom a message is directed. Therefore, a crucial element is understanding who the speaker and who the addressee/es are. This ability is even more crucial for social robots in situations that go beyond the mere dyadic interactions. For instance, understanding others' addressee in a group dynamics could help robots to discern implicitly expressed robot-directed commands \cite{Bakx2003}, the social dynamics and roles in multiparty interactions \cite{Strohkorb2015}, and the correct meaning of sentences that contains deictic expressions (you, he, she, they...) \cite{Gold2009Deixis}.


The scope of the present study is to implement a model for Addressee Estimation to enhance Human-Robot Interaction (HRI). Addressee Estimation is the capability to detect the addressee of a user's utterance \cite{Skantze2020}. Endowed with such skill, robots would be able to estimate \emph{the addressee: to whom an agent is addressing its message} \cite{JakobsonLinguisticsandPoetics}. Behavioral studies demonstrated that the human expression of communicative aspects related to Addressee Estimation (i.e., turn yielding and turn taking) involves verbal, para-verbal and non-verbal channels \cite{Skantze2020}. Specifically it was proven that, for Addressee Estimation, the speaker's bodily cues, such as gaze and gestures, allow listeners to better understand the speaker’s intentions \cite{Auer2018gaze, Ishii2016}. Our approach in developing our Addressee Estimation model for social robots was inspired by these findings. Hence, we conceive Addressee Estimation as \emph{the ability to understand an utterance's addressee by interpreting and exploiting non-verbal/bodily cues from the speaker.} 

\section{Related Works}
Previous studies have often dealt with the problem of Addressee Estimation by using a multimodal approach. Jovanovic et al. \cite{Jovanovic2006} used an ad-hoc retrieved dataset, gathered on meetings of groups of 4 humans, to train a Bayesian Network and Na\"{\i}ve Bayes Classifiers with contextual, lexical, and gaze features and solve the task as a classification of whom, among the four agents (or the group), was the addressee of each utterance. 


Using the AMI corpus \cite{Carletta2005}, containing data from 100 hours of meetings, Op den Akker et al. \cite{opdenakker2009} treated Addressee Estimation as a binary problem from the perspective of each agent and trained different classifiers with data about the speakers' focus of attention, dialogue acts, and contextual information. Using the same dataset, Malik et al. \cite{Malik2019} selected several features (textual, contextual, and focus of attention) to classify the role of the addressee instead of their identity. In this way, they overcame some limitations in terms of reducing Addressee Estimation to a mere binary problem. In a later study, the same authors used similar features on the MULTISIMO Corpus, a multiparty multimodal dataset involving meetings of 3 participants \cite{Malik2021}. In this work, the authors trained different machine learning and deep learning algorithms to improve their previous results and develop a real-time Addressee Estimation model, without information about previous addressees.

In embodied artificial agents, Addressee Estimation models have started to be developed to go beyond the dyadic and robot-centric structure of HRI \cite{johansson2015opportunities, Sheikhi2013} and enable robots to interact in ecological scenarios \cite{Horiguchi2019}. 
However, research on Addressee Estimation, which directly involved artificial conversational systems and robots, mostly solved the problem as a binary classification. Bakx et al. \cite{Bakx2003} conducted multiparty experiments with two humans at an information kiosk. By recording participants with an external camera, they used a rule-based approach to classify whether the participant at the information kiosk was addressing the system or the human partner, given its focus of attention and the length of the utterance. Operating with the same scenario, Turnhout et al. \cite{Turnhout2005} trained a Na\"{\i}ve Bayes Classifier to solve the same task. Katzenmaier et al. \cite{Katzenmaier2004} designed a multiparty interaction with two humans (host and guest) and a simulated robot. They approached the task as a binary classification (host speaking either to the robot or to the guest) using visual data (automatically extracted head pose) and speech data. In a multiparty HRI scenario, Richter et al. \cite{Richter2016smart} opted for a rule-based model taking as input the human's lips movement and the mutual gaze between the human and the robot to understand if an utterance was addressed to the robot or not. After a dataset collection of multi-user human-virtual agent interaction, Huang et al. \cite{Huang2011} trained an SVM classifier for a binary classification (robot addressed or human partner addressed) by giving as input several features related to prosody, utterance length, and head direction and equipped the virtual agent with a model for real-time Addressee Estimation. The work of Sheikhi et al. \cite{Sheikhi2013} relied on the role contextual information plays in Addressee Estimation. The authors used context about the utterance, the agents involved in the interaction, and the objects of interest in the environment to extract information about the speaker's and the human partner's visual focus of attention to train a model to predict the addressee of each utterance in a binary classification task.

Addressee Estimation has also been connected to other social communication problems. Johansson et al. \cite{johansson2015opportunities} combined turn-taking and addressee detection and used only automatically extracted features to solve the task as binary classification and as a linear regression problem (gradual opportunity to take the turn).  Horiguchi et al. \cite{Horiguchi2019} trained a LSTM neural network combined with Logistic Regression with features related to the speaker's face, audio and text to solve the problem of response obligation detection. Romeo et al. \cite{Romeo2019} implemented a CNN model on a Pepper robot to endow it with the ability to predict the other agent's intention to interact using only visual information.

To develop an Addressee Estimation model that could be deployed in social robots, the approach of our work is to design and train a hybrid deep-neural network (CNN+LSTM) to interpret the non-verbal behavior of the speaker as a cue to localize the addressee in the HRI space. The idea is to provide the robot with the ability to read the speaker's intentions only with the visual information on their behaviour. This work considers Addressee Estimation with the idea of enabling multiparty natural HRI. For this reason, with respect to previous models implemented on robots, our work answers to three needs: going beyond the mere binary prediction (robot being addressed or not), using only data from the robot's sensors, and taking into account more realistic and ecological scenarios. 

\section{Methods}
The core objective of this study is to develop a model for Addressee Estimation in terms of addressee localisation. We approached the task as a three-class classification of the addressee position starting from the speaker's non-verbal behaviour and we designed a deep neural network composed of two parts: a CNN and an LSTM network. As Skatnze \cite{Skantze2020} showed, bodily information can be crucial to estimate the addressee of an utterance. Based on such evidences, the model developed in this work leverage two types of visual data collected from the speaker: 1) Images of the speakers' face, to retrieve information about their head direction and visual focus. 2) Vectors representing the speakers' body pose, as a proxy of their focus of attention but also as a way to encapsulate the body language and gestures.

\subsection{The dataset} \label{ape_dataset}
We decided to use the Vernissage Corpus \cite{Vernissage, Vernissage2013} to train and test our model for Addressee Estimation. Vernissage is a synchronised multimodal corpus of multiparty interactions in which two humans converse with a Nao robot (Aldebaran, United Robotics Group). During the interaction, the robot asks participants to present themselves to the group, shows them some paintings on the walls of the room, and asks them some questions about the paintings. Participants are given time to discuss among themselves before answering the robot, and are left free to comment with their peer on the situation. The aim of the corpus was to leave participants as free as possible during the interaction. All recordings take place in the same room, where participants are not required to keep a specific absolute position, although most of the time, because of the configuration of the room and of the interaction, they are standing in front of Nao and hold a relative position to each other: one on the left, the other on the right side (see Figure \ref{fig:Dataset} for some examples). Dialogues about the paintings on the walls ensure the presence of a more complex and natural interaction. More specifically, triadic interactions (two agents and a target object) occur when the speaker talks to the addressee about something in the environment, might it be an object or another agent. This situation, which is typical of social interactions, modifies the gaze pattern and the body posture of both agents so that they not only look at each other but also at the target object \cite{Skantze2020}. In the Vernissage Corpus, participants focus their attention on the paintings and even get closer to see them better when describing them. Therefore, although the dialogue is controlled by the questions of the robot, the scenario grants sufficient flexibility and spontaneity to human behavior, as it normally happens in ecological scenarios.

\begin{figure}[t!]
\centerline{\includegraphics[width=1\columnwidth]{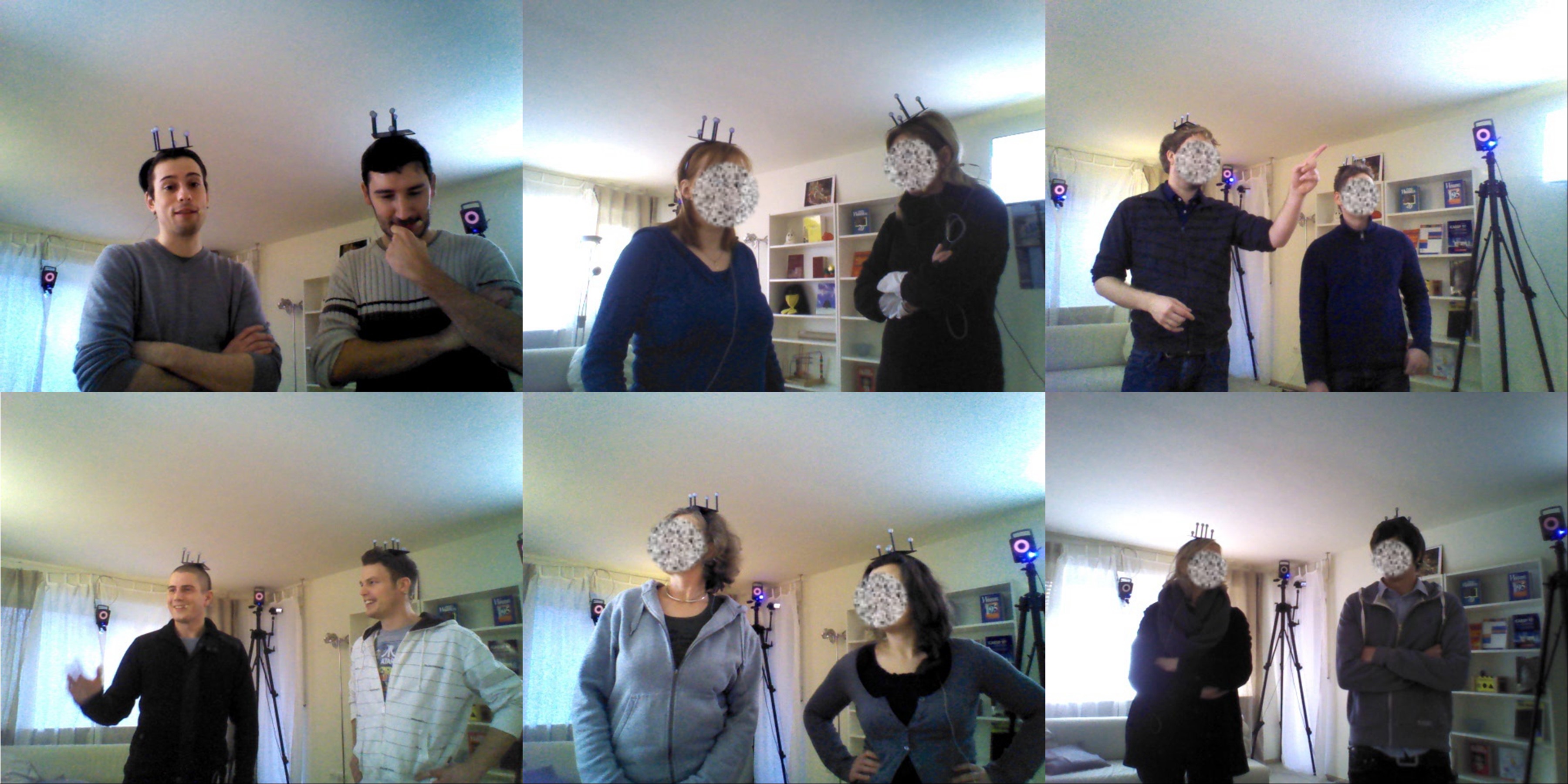}}
\caption{\textbf{Illustrative frames from Vernissage Dataset.} Examples of multiparty HRI data recorded from the Nao robot's cameras. Some pictures show blurred faces for privacy reasons.}
\label{fig:Dataset}
\end{figure}

\begin{figure}[b!]
\centerline{\includegraphics[width=1\columnwidth]{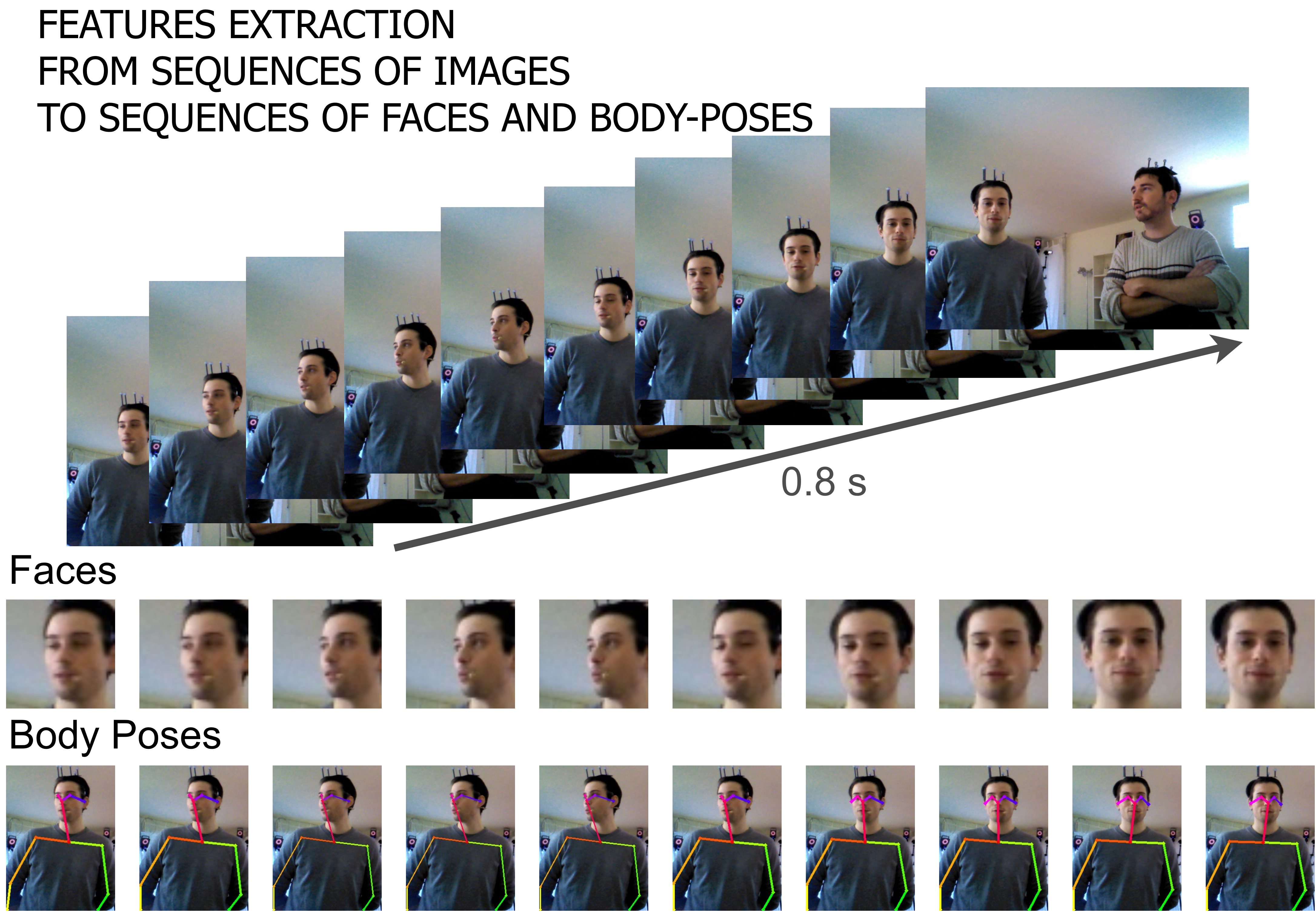}}
\caption{\textbf{Illustration of a sequence.} Aggregation of frames in a sequence of 0.8s and extraction of body poses and face images}
\label{fig:sequence}
\end{figure}

Yet another positive feature of the Vernissage Corpus is that the authors recorded the interactions from the cameras of the Nao robot, a single camera installed in the head of the robo with a resolution of 640x480 pixels at a frequency of about 15 fps (mean), and YUV422 color mode. 
The idea guiding the development of our model was to be able to carry out Addressee Estimation in unconstrained interaction settings. With this objective in mind, we chose to focus on data that could be automatically extracted through the robot's sensors without the need for any external device. This way, the model could be trained with data recorded from the ego-centric perspective of a robot, extremely important for portability. 
As a side effect, data result noisier because of the movements made by the robot (e.g., nodding or turning its head). However, this could also represent an advantage while training a model suitable for ecological scenarios.

The Vernissage Corpus is manually annotated to have a Ground Truth about addressees. More specifically, the addressee is annotated as the target person of a speech utterance. 
Five different labels are used: ``ROBOT'', ``RIGHT'', ``LEFT'', ``GROUP'', and ``NOLABEL''. ``RIGHT'' and ``LEFT'' refer to the person at the right and the left of the robot, ``GROUP'' means that both the robot and the other agent are addressed, whereas ``NOLABEL'' indicates a silent time interval. 

\subsection{Features selection and pre-processing data pipeline} \label{ape_preprocessing}
The data chunk on which our neural network for Addressee Estimation has been trained is a sequence of 10 frames of face images and vectors representing body poses. The process to obtain the chunks from the Vernissage Corpus involved the following five steps: 

\paragraph{Division in utterances}
The dataset comprises recordings of 10 interactions between two humans (different for each interaction) and the Nao robot. Video clips were trimmed according to the speech detection annotations and extracted from the recordings of the 10 interactions, leaving out the frames labeled ``silence''.  Therefore, utterances were considered as the time intervals in which an agent continuously speaks without being interrupted by silence pauses longer than 0.08s. 

\begin{figure*}[htb!]
\centerline{\includegraphics[width=1\textwidth]{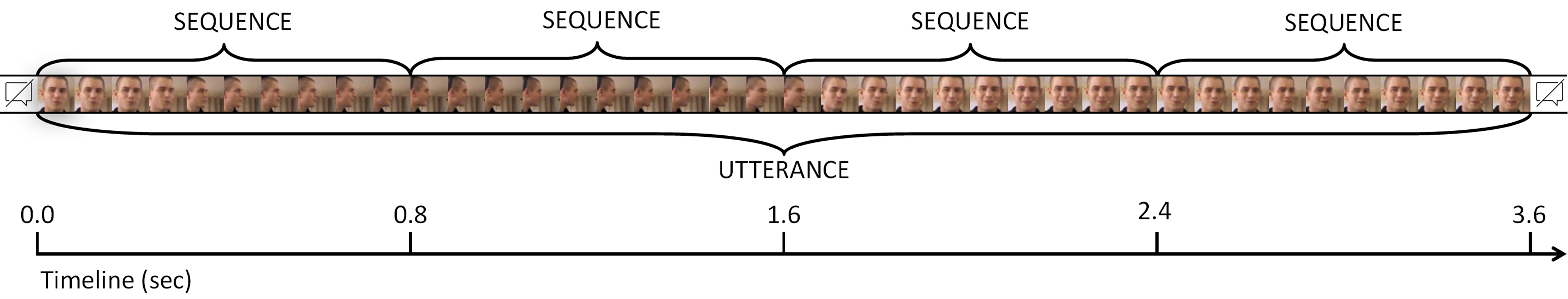}}
\caption{\textbf{Illustration of an utterance.} The utterance is partitioned into sequences of 0.8s. Utterances were defined as speech intervals addressed to the same addressee and delimited by silence. Each utterance comprised at least one sequence.}
\label{fig:sequence_utterance}
\end{figure*}

\paragraph{Extraction of body poses and face images}
Firstly, utterances were divided into frames of 0.08s. Then, using OpenPose \cite{Openpose2019}, vectors of 2D coordinates for body poses were extracted for each frame for all participants. The OpenPose COCO body format was adopted, which predicts the x and y coordinates for 18 key points of each person (5 for the head, 3 for each limb, and 1 for the torso). Coordinates ranged from -1 to 1 for both axes. Given the poses, from the five coordinates of each person's head-key points, a square-size cropped face image was obtained by the original frame, resized at 50x50px. Body poses and face images were labeled as ``speaker'' or ``other''. During the interactions in the Vernissage Corpus, both human participants play the role of ``speaker'' and ``other''. Hence, since the corpus comprises 10 interactions, each with two possible speakers, 20 instances of speakers were finally available.

\paragraph{Aggregation in sequences}
Sequences, thereby, resulted being 0.8s portions of the utterances, consisting of 10 body poses and 10 face images. Since the objective was to train a network with only data from the speaker, the speaker's sequences were saved separately from those belonging to the other participant. All sequences were annotated with the addressee (``ROBOT'', ``LEFT'', ``RIGHT'', ``GROUP''). Figure \ref{fig:sequence} shows an illustrative sequence, whereas Figure \ref{fig:sequence_utterance} shows the difference between sequences and utterances.

\paragraph{Data augmentation}
We tackle the implementation of Addressee Estimation as the classification of the addressee's position with respect to the speaker and from the ego-centric perspective of the robot. Therefore, we selected three labels among the ones already annotated in the original corpus, as the addressee's position could be classified as ``LEFT'', in case the addressee was at the left of the speaker (from the robot’s perspective), ``RIGHT'', in case the addressee was at the speaker's right, or ``ROBOT'', in case the addressee was the robot. 
The interaction scenario in Vernissage, with the Nao robot asking questions and managing the interaction, caused an imbalanced representation of classes in the dataset, with a prevalence of sequences labeled as ``ROBOT'' (addressed to the robot). With the twofold objective of augmenting the dataset and balancing the number of sequences, all frames labeled as ``LEFT'' and ``RIGHT'' (and accordingly, the body poses and the face images extracted by them) were flipped, and their label inverted. As a consequence, ``LEFT'' and ``RIGHT'' data were doubled, and the resulting dataset was composed of 18190 speaker's face images and body poses partitioned in 1819 sequences: 529 for ``ROBOT'', 645 for ``LEFT'', and 645 for ``RIGHT''.

\paragraph{Body pose shifting}
As it appears from Figure \ref{fig:Dataset}, participants at the left of the robot never spoke toward a left addressee, and participants at the right never did it toward a right one. Therefore, even though participants could mildly move, the coordinates of their bodies could bias the prediction of their addressee. To overcome this issue, for each sequence, the 10 body poses were shifted along the x-axis of a random measure ranging from the two extremes of the image.

\subsection{Architecture Design} \label{ape_nn_design}

Our Addressee Estimation model is a CNN + LSTM hybrid architecture that extracts features through convolutions and then supports learning temporal sequential patterns through LSTM cells. Previous works related to human activity recognition combined CNNs with an LSTM final layer. For instance, Subramaniam et al. \cite{Subramaniam2016} used this combination to train a model for classifying first impressions of personality. Romeo et al. \cite{Romeo2021} used a similar architecture to predict apparent personality from body language cues for HRI. Moreover, Ullah et al. \cite{Ullah2018} integrated convolutional and LSTM layers for action recognition from videos, while Nakisa et al. \cite{Nakisa2020} developed a multimodal neural network with convolutional and LSTM layers for emotion recognition through physiological signals. 

Our model is also designed to exploit and integrate both visual input modalities: the face images and the body pose vectors. Face images and body pose vectors pass independently in two parallel streams of convolutional layers. Consequently, the two embeddings received as output are concatenated before the LSTM layer. In this way, features are extracted separately in convolutions and then combined at a higher level of abstraction. This was inspired by a gradual fusion of modalities at an intermediate level of the network, which has been demonstrated beneficial \cite{Ramachandram2018, Stahlschmidt2022}, and by the training on joint representations of temporal sequences as, for instance, in Nakisa et al. \cite{Nakisa2020}, which proved fusing streams between the convolutional and LSTM layers being beneficial rather than a late fusion after temporal training. 

Therefore, our Addressee Estimation model implements the intermediate-fusion approach and consists of two blocks, each including two 2D convolutional layers (the second followed by a LeakyReLU activation function) and one max-pooling layer. The two convolutional blocks are followed by two fully connected layers (the first followed by a LeakyReLU activation function) providing the embeddings of the input modalities to be concatenated. 
The fusion of the two streams is carried out as a simple concatenation, with the body pose embeddings repeated 29 times so as to balance the information in the final embedding. The sequence of the 10 fused embeddings is then passed through the LSTM layer. Eventually, after two final fully connected layers (the first followed by a LeakyReLU activation function), the output is given by a LogSoftMax layer, as shown in Figure \ref{fig:nn} and described in Table \ref{tab:nn_composition_late}.
\begin{figure*}[t!]
\centerline{\includegraphics[width=1\textwidth]{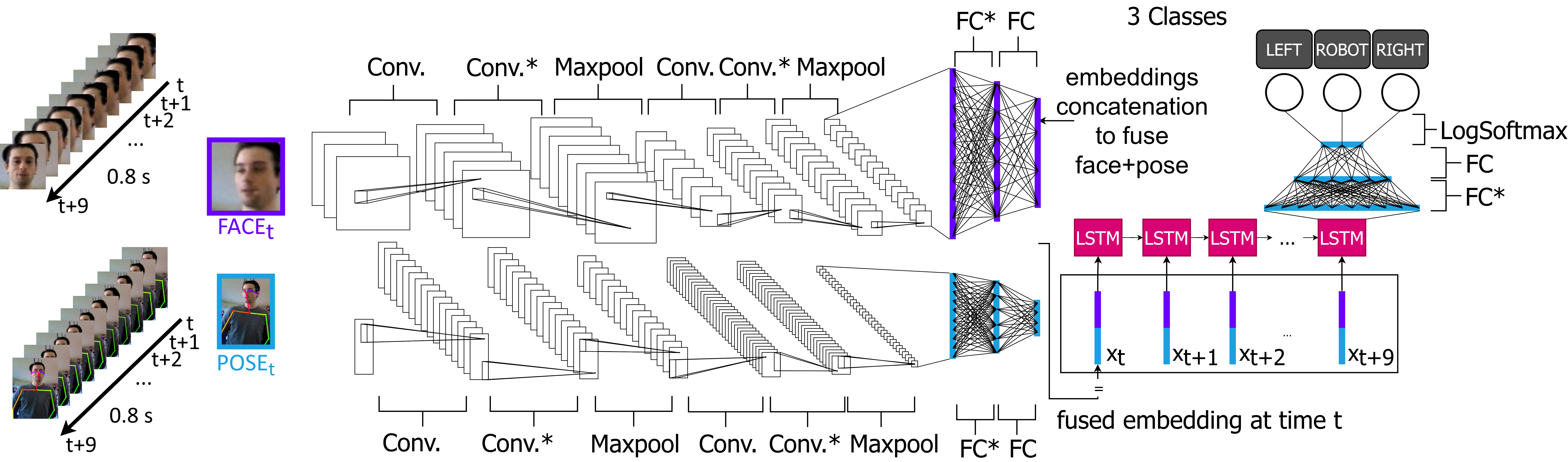}}
\caption{\textbf{Illustration of the Deep Neural Network for Addressee Estimation employing an intermediate fusion approach (Exp. 1a).} Face images and body pose vectors are passed separately to two blocks of convolution, each including two 2D convolutional and one max-pooling layers. Then, the two embeddings resulting from fully connected layers are concatenated and sequences of 10 fused embeddings are passed to the LSTM layer. The output is provided after two others fully connected layers and a LogSoftMax layer. * represents LeakyReLU activation function.}
\label{fig:nn}
\end{figure*}

\subsection{Training Procedure and Experiment Lineup} \label{ape_training_procedure}
To train and test our model, a 10-fold cross-validation was established. In this way, the prediction of the classes (``LEFT'', ``RIGHT'', ``ROBOT'') could be evaluated based on the average performance of the model when trained and tested on different sets of data. To create the 10 different train and test sets, the dataset of sequences derived from the pre-processing of the Vernissage Corpus was partitioned along the 10 multiparty interactions of the original corpus. Each interaction comprised two agents for a total of 20 speakers. The ratio to create the train sets was 9:10, with the remaining 1:10 for the test set. Accordingly, each train set included all the face image and body pose sequences of 18 participants, whereas the test set the ones of the remaining 2 participants. From the train sets, 90 sequences (30 for each class) were randomly extracted, removed, and used for the validation phase during the training in order to check the trend of the loss function.

As a first experiment (Exp 1.a), the addressee localisation model was thus trained 10 times, one for each train set, and evaluated on as many test sets. The model was fed with temporal sequences of data in mini-batches of 10 sequences. Each sequence included 10 face images and 10 body pose vectors. The convolutional section of the network for face images was trained by employing Stochastic Gradient Descent (SGD) optimizer whereas the one for body pose vectors and the LSTM section were trained using the Adam algorithm \cite{AdamMethod}. Cross entropy was used as the criterion to compute the loss function. The model was trained for 50 epochs with a learning rate of 1e-3, with a decay of 0.1 (multiplicative factor) after 40 epochs. To prevent overfitting, early stopping was implemented to stop the training after 10 trials in which the loss function of the evaluation phase increased. The model was implemented using PyTorch 1.12 (Python version 3.8), whereas the training was carried out through an NVIDIA Quadro RTX 5000 with 16 GB of RAM\footnote{the code is available at the following link \\ https://gitlab.iit.it/cognitiveInteraction/addressee\_estimation\_ijcnn23.git}.

In order to test the developed intermediate-fusion model, and validate the results from Exp 1.a, we conducted three additional experiments:
\begin{itemize}
    \item Exp. 1.b: our original model was modified to a late-fusion approach, to be compared to our intermediate-fusion model
    \item Exp. 1.c: our model was trained on face images only, to verify whether the two input modalities were equally necessary
    \item Exp. 1.d: single modality: our model was trained on body pose vectors only, to verify whether the two input modalities were equally necessary
\end{itemize}

In Exp. 1.b, we trained the model using both visual input modalities but, differently from the original model, we combined them after the LSTM layer. The fusion of the two streams was carried out as a concatenation of the two embeddings resulting from the LSTM layer. In this way, for each sequence, one fused embedding was passed through the two final fully connected layers (the first followed by a LeakyReLU activation function) and the LogSoftMax layer. In Exp. 1.c and 1.d, we trained and tested two mono-stream models, the former with face images, the latter with body pose vectors. For each modality, the architecture of the neural network matched the late-fusion approach model except for the last fully connected layers, which were designed for mono-modality embeddings. Table \ref{tab:nn_composition_late} provides a description for the four architectures of Exp. 1.a-b-c-d.

Finally, we carried out a last experiment (Exp. 2) to compare our results from Exp 1.a with the state-of-the-art model used on Vernissage as a binary classification, predicting whether either the robot or another user was the addressee of an utterance \cite{Sheikhi2013}. Therefore, we trained and tested our intermediate-fusion binary model to identify whether the robot was addressed by the speaker. For this experiment, we used the same architecture employed in Exp. 1.a, except for the last two layers (last fully connected and LogSoftMax), changed to give binary predictions. For this reason, the previous three-class data labeling of the addressee's position (``LEFT'', ``ROBOT'', ``RIGHT'') was replaced by a binary one: ``NOT-ADDRESSED'' (including data referred to as ``LEFT'' and ``RIGHT'' ) and ``ADDRESSED'' (including data referred to as ``ROBOT'' and ``GROUP'' in the original Vernissage labeling system).

\begin{table}
\centering
\caption{\textbf{Description of neural networks in the four three-classes experiments (Exp. 1.a-b-c-d).}}
\label{tab:nn_composition_late}
\begin{tabular}{|l|ll|ll|} 
\toprule
\multicolumn{1}{l|}{}         & \multicolumn{2}{l|}{\textbf{FACE IMAGE}}            & \multicolumn{2}{l|}{\textbf{BODY POSE VECTOR}}       \\ 
\hline
\textbf{Layers}               & \textbf{Input}    & \textbf{Param.}                 & \textbf{Input}  & \textbf{Param.}                    \\ 
\hline
\multicolumn{5}{|l|}{\textit{Layers common to all models (convolutional part of the nn)}}                                                  \\ 
\hline
Conv                          & {[}100,3,160,160] & k=7,s=1                         & {[}100,1,18,3]  & k=(3,1),s=1                        \\ 
\cline{1-1}
Conv*                         & {[}100,6,154,154] & k=5,s=1                         & {[}100,16,16,3] & k=(3,1),s=1                        \\ 
\cline{1-1}
MPool                         & {[}100,8,150,150] & k=2,s=2                         & {[}100,16,14,3] & k=(2,1),s=2,1                      \\ 
\cline{1-1}
Conv                          & {[}100,8,75,75]   & k=5,s=1                         & {[}100,16,7,3]  & k=(3,1),s=1                        \\ 
\cline{1-1}
Conv*                         & {[}100,12,71,71]  & k=3,s=1                         & {[}100,32,5,3]  & k=(3,1),s=1                        \\ 
\cline{1-1}
MPool                         & {[}100,16,69,69]  & k=2,s=2                         & {[}100,32,3,3]  & k=2,s=2                            \\ 
\cline{1-1}
Flatten                       & {[}100,16,34,34]  &                                 & {[}100,32,1,1]  &                                    \\ 
\cline{1-1}
FC*                           & {[}100,18496]     &                                 & {[}100,32]      &                                    \\ 
\cline{1-1}
FC                            & {[}100,4624]      &                                 & {[}100,24]      &                                    \\ 
\cline{1-1}
\multicolumn{1}{:l:}{...}     & ...               & \multicolumn{1}{l:}{}           & ...             & \multicolumn{1}{l:}{}              \\ 
\hdashline
\multicolumn{1}{l}{}          &                   & \multicolumn{1}{l}{}            &                 & \multicolumn{1}{l}{}               \\
\multicolumn{1}{l}{}          &                   & \multicolumn{1}{l}{}            &                 & \multicolumn{1}{l}{}               \\ 
\hdashline
\multicolumn{5}{:l:}{\textit{Layers after convolution in Exp. 1.a.: Intermediate fusion model}}                                            \\ 
\hdashline
\multicolumn{1}{:l:}{Concat.} & {[}100,578]       & \multicolumn{1}{l:}{}           & \multicolumn{2}{l:}{{[}100,20] x 29 times}           \\ 
\cline{1-1}
FC*                           & {[}10,256]        & \multicolumn{1}{l}{}            &                 &                                    \\ 
\cline{1-1}
FC                            & {[}10,128]        & \multicolumn{1}{l}{}            &                 &                                    \\ 
\cline{1-1}
LSoftm.                       & {[}10,3]          & \multicolumn{1}{l}{}            &                 &                                    \\ 
\hline\hline
\multicolumn{1}{l}{}          &                   & \multicolumn{1}{l}{}            &                 & \multicolumn{1}{l}{}               \\
\multicolumn{1}{l}{}          &                   & \multicolumn{1}{l}{}            &                 & \multicolumn{1}{l}{}               \\ 
\hdashline
\multicolumn{5}{:l:}{\textit{Layers after convolution in Exp. 1.b.: Late fusion model }}                                                   \\ 
\hdashline
\multicolumn{1}{:l:}{LSTM}    & \multicolumn{2}{l:}{{[}10,10,578] ~~ h\_dim=512}    & \multicolumn{2}{l}{{[}10,10,20] ~~ h\_dim=256}       \\ 
\cline{1-1}
FC                            & {[}10,512]        &                                 & {[}10,256]      &                                    \\ 
\cline{1-1}
Concat.                       & {[}10,128]        &                                 & {[}10,128]      &                                    \\ 
\cline{1-1}
FC*                           & {[}10,256]        & \multicolumn{1}{l}{}            &                 &                                    \\ 
\cline{1-1}
FC                            & {[}10,128]        & \multicolumn{1}{l}{}            &                 &                                    \\ 
\cline{1-1}
LSoftm.                       & {[}10,3]          & \multicolumn{1}{l}{}            &                 &                                    \\ 
\hline\hline
\multicolumn{1}{l}{}          &                   & \multicolumn{1}{l}{}            &                 & \multicolumn{1}{l}{}               \\
\multicolumn{1}{l}{}          &                   & \multicolumn{1}{l}{}            &                 & \multicolumn{1}{l}{}               \\ 
\hdashline
\multicolumn{5}{:l:}{\textit{Layers after convolution in Exp. 1.c (Face model) and 1.d (Pose model)}}                                      \\ 
\hdashline
\multicolumn{1}{:l:}{}        & \multicolumn{2}{l:}{\textit{Exp. 1.c.: Face Model}} & \multicolumn{2}{l:}{\textit{Exp. 1.d.: Pose Model}}  \\ 
\hdashline
\multicolumn{1}{:l:}{LSTM}    & \multicolumn{2}{l:}{{[}10,10,578] ~ h\_dim=512}     & \multicolumn{2}{l}{{[}10,10,20] ~~ h\_dim=256}       \\ 
\cline{1-1}
FC*                           & {[}10,256]        &                                 & {[}10,256]      &                                    \\ 
\cline{1-1}
FC                            & {[}10,128]        &                                 & {[}10,128]      &                                    \\ 
\cline{1-1}
LSoftm.                       & {[}10,3]          &                                 & {[}10,3]        &                                    \\ 
\hline\hline
\multicolumn{5}{l}{\begin{tabular}[c]{@{}l@{}}k: kernel size; s: stride; h\_dim: hidden layer dimension\\ *: + LeakyReLU\end{tabular}}     \\
\bottomrule
\end{tabular}
\end{table}

\subsection{Evaluation Metrics} \label{ape_evaluation_metrics}
To evaluate our model Precision, Recall, and F1-score were computed for each class and expressed as a percentage. 
Results for each class were subsequently weighted for the number of samples of each class and averaged to provide a performance of each model in terms of Weighted F1-score. Eventually, the results of the 10 testing from the 10-fold cross-validation were averaged to obtain a final estimate of the model's performances. 

In all experiments, the training and testing of the model were first achieved by keeping the 10-frames sequences as data chunks. However, since sequences were extracted from utterances, the same metrics could be used to verify the model's performance in predicting the addressee of an entire utterance. The utterance classification was computed by averaging the predictions of all the sequences belonging to that utterance, weighted for the prediction score provided by the LogSoftMax layer.

For Exp. 2, Precision, Recall (Sensitivity), F1-score, and Specificity (true negative rate) were calculated considering the positive prediction as ``the robot is addresse'' and expressed as a percentage. 
In addition to these parameters, for a further measure of the model's performance, an overall-F1-score of the two classes (``ADDRESSED'' vs ``NOT-ADDRESSED'') was computed as in the three-class model. 

\section{Results} \label{ape_results}


In a first set of experiments (Exp. 1.a-b-c-d), we trained the Addressee Estimation model on a three-class classification task. 
The four experiments (a-b-c-d) were designed to verify the best approach for the Addressee Estimation model: either multi-feature (and more specifically an intermediate (1.a) of late-fusion (1.b) approach) or mono-feature (and more specifically using the face (1.c) or body pose (1.d) information).

\begin{table}
\centering
\caption{\textbf{Performance of the Addressee Estimation model.} Results of the 10-fold cross-validation experiments (Exp. 1.a-b-c-d) are provided in terms of mean and standard deviation of weighted F1-score.}
\label{tab:results_nn}
\begin{tabular}{|l|ll|ll|ll|} 
\cline{2-7}
\multicolumn{1}{l|}{}                                                  & \multicolumn{2}{l|}{\begin{tabular}[c]{@{}l@{}}considering\\ sequences\end{tabular}} & \multicolumn{2}{l|}{\begin{tabular}[c]{@{}l@{}}considering\\utterances\end{tabular}} & \multicolumn{2}{l|}{\begin{tabular}[c]{@{}l@{}}considering \\1st seq. \\of each utt.\end{tabular}}  \\ 
\hline
Model                                                                  & avg   & std                                                                          & avg   & std                                                                          & avg   & std                                                                                         \\ 
\hline
\begin{tabular}[c]{@{}l@{}}Exp. 1.a \\Intermediate Fusion\end{tabular} & 75.01 & 8.60                                                                         & 76.48 & 8.42                                                                         & 74.15 & 9.19                                                                                        \\ 
\hline
\begin{tabular}[c]{@{}l@{}}Exp. 1.b\\Late Fusion\end{tabular}          & 73.18 & 7.57                                                                         & 74.19 & 7.97                                                                         & 71.88 & 5.03                                                                                        \\ 
\hline
\begin{tabular}[c]{@{}l@{}}Exp. 1.c\\Only Face\end{tabular}            & 72.83 & 5.86                                                                         & 73.22 & 6.93                                                                         & 72.07 & 8.89                                                                                        \\ 
\hline
\begin{tabular}[c]{@{}l@{}}Exp. 1.d\\Only Body Pose\end{tabular}       & 72.60 & 6.75                                                                         & 71.05 & 7.76                                                                         & 70.77 & 8.84                                                                                        \\
\hline
\end{tabular}
\end{table}

In each experiment, the model was first tested on single sequences of 0.8s, hence without combining them in utterances (Figure \ref{fig:sequence_utterance} illustrates the difference between sequences and utterances). In terms of weighted F1-score, in Exp. 1.a, the model achieved 75.01\% , whereas in Exp. 1.b, 1.c, and 1.d the performance was, respectively, 73.18\%, 72.83\% and 72.61\% (see Table \ref{tab:results_nn} and Figure \ref{fig:F1-score}).

However, Addressee Estimation is defined as the ability to understand the addressee of an utterance. Accordingly, we tested the model considering sequences of the same utterance together, because each utterance might comprise several sequences. In this case, in Exp. 1.a the weighted F1-score increased up to 76.48\%, whereas in Exp. 1.b, 1.c and 1.d, the performance measured, respectively, 74.19\%, 73.22\% and 71.05\%.

An additional score was computed only focusing on the first sequence of each utterance, which means measuring the model's performance in providing a correct prediction at 0.8s from the beginning of the utterance. Considering the limited amount of time, the model achieved a weighted F1-score of 74.15\% in Exp. 1.a, whereas in 1.b, 1.c, and 1.d, it was, respectively, 71.88\%, 72.07\% and 70.77\%.

\begin{figure*}[!ht] 
\centerline{\includegraphics[width=\textwidth]{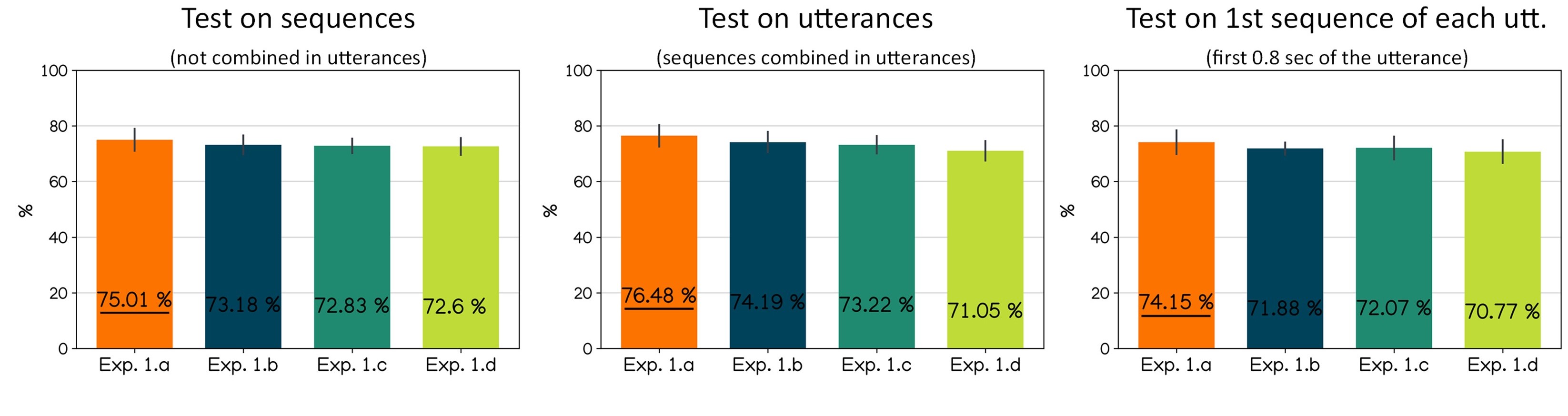}}
\caption{\textbf{Bar plots reporting performance of Addressee Estimation model in the four 3-class experiments.} Results of the 10-fold cross-validation experiments (Exp. 1.a-b-c-d) are provided in terms of mean and standard deviation (error bar) of weighted F1-scores. 
On the y-axis the performance score is expressed in \%.}
\label{fig:F1-score}
\end{figure*}

\begin{figure}[!ht] 
\centerline{\includegraphics[width=1\columnwidth]{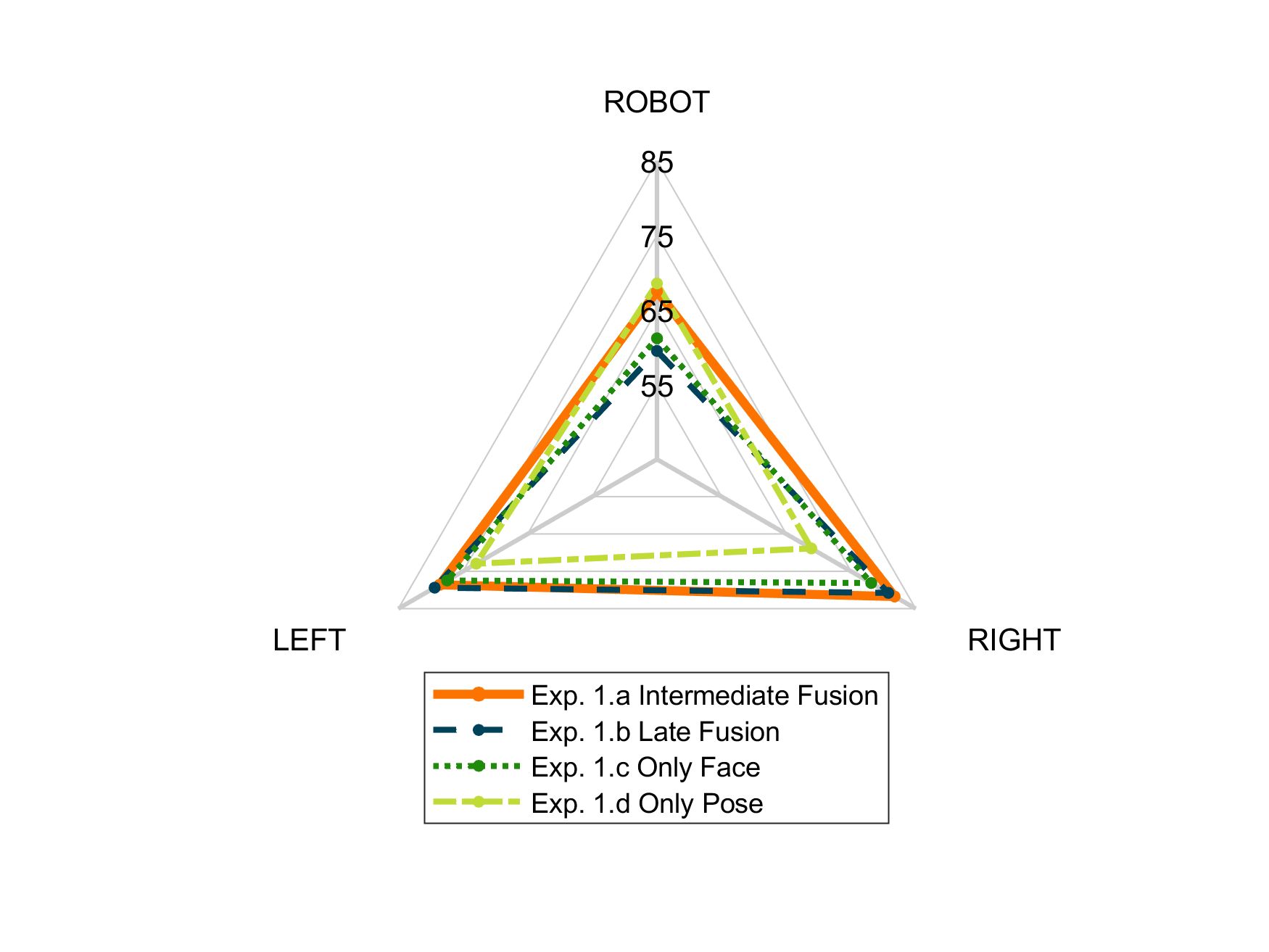}}
\caption{\textbf{Spider plot reporting performance of Addressee Estimation model for each class.} Results of the 10-fold cross-validation experiments (Exp. 1.a-b-c-d) are provided in terms of weighted F1-score (means) for each of the 3 classes and expressed in \%.}
\label{fig:class_performances}
\end{figure}

Precision, Recall, and F1-score have been computed for each class (``LEFT'', ``ROBOT'', ``RIGHT'') to observe if the model performances resulting from the 4 experiments (1.a-b-c-s) were equally distributed among classes. Results are reported in Table \ref{tab:results_classes_nn} and visually displayed in Figure \ref{fig:class_performances}.

\begin{table}
\centering
\caption{\textbf{Performance of Addressee Estimation models for each class.} Results of the 10-fold cross-validation experiments experiments (Exp. 1.a-b-c-d) are computed considering sequences not combined in utterances}
\label{tab:results_classes_nn}
\begin{tabular}{|l|l|l|l|l|} 
\hline
Model    & Class & recall & precision & F1-score  \\ 
\hline
~        & LEFT  & 80.72  & 79.07     & 78.61     \\
Exp. 1.a & ROBOT & 70.09  & 72.05     & 67.52     \\
         & RIGHT & 81.76  & 84.34     & 81.75     \\ 
\hline
         & LEFT  & 83.34  & 77.28     & 79.38     \\
Exp. 1.b & ROBOT & 54.91  & 72.63     & 59.51     \\
         & RIGHT & 89.22  & 75.43     & 80.78     \\ 
\hline
         & LEFT  & 78.43  & 77.58     & 77.46     \\
Exp. 1.c & ROBOT & 62.83  & 68.59     & 61.21     \\
         & RIGHT & 80.73  & 78.11     & 78.13     \\ 
\hline
         & LEFT  & 71.99  & 76.84     & 72.95     \\
Exp. 1.d & ROBOT & 75.16  & 65.37     & 68.56     \\
         & RIGHT & 67.11  & 73.19     & 68.87     \\
\hline
\end{tabular}
\end{table}


Our model was trained on sequences of 0.8s but utterances can be composed of multiple sequences. For such utterances, fresh updated predictions can be released every 0.8s, computed by averaging the prediction of each sequence. Therefore, we could check the performance of utterances' Addressee Estimation as time passes. Figure \ref{fig:incremental_results} displays the trend of such performances in terms of weighted F1-score in the four experiments (1.a-b-c-d). For what concerned Exp. 1.a, the weighted F1-score was 74.15\% at 0.8s, 76.48\% at 1.6s, 76.5\% at 2.4s and 79.8\% after 2.4s. At the same time intervals, in Exp 1.b, the performance measured respectively 71.88\%, 72.03\%, 75.22\%, and 77.22\%; in Exp. 1.c, 72.07\%, 75.04\%, 77.26\%, and 78.25\%; and in Exp 1.d, 70.77\%, 70.23\%, 70.44\%, and 70.49\%. 

In Exp. 2, we trained our model to solve Addressee Estimation as a binary classification task in the shape of the robot being addressed or not  (``ADDRESSED'' vs ``NOT-ADDRESSED'').
Considering the test on single sequences, our repurposed model's Recall to the affirmative answer (robot addressed) was 73.78\%, whereas Precision was 74.23\%, and F1-score 72.73\%. Sensitivity achieved 80.7\%. Additionally, the general performance of the model, as measured by the overall-F1-score was 77.36\% if measured on single sequences, 79.7\% considering utterances, and 79.97\% considering the first 0.8s of each utterance. The performance of the model in Sheikhi et al. \cite{Sheikhi2013} was 76.3\% utterances correctly predicted employing a measure of the speaker's visual focus of attention automatically computed as input. 

\begin{figure}[!ht] 
\centerline{\includegraphics[width=1\columnwidth]{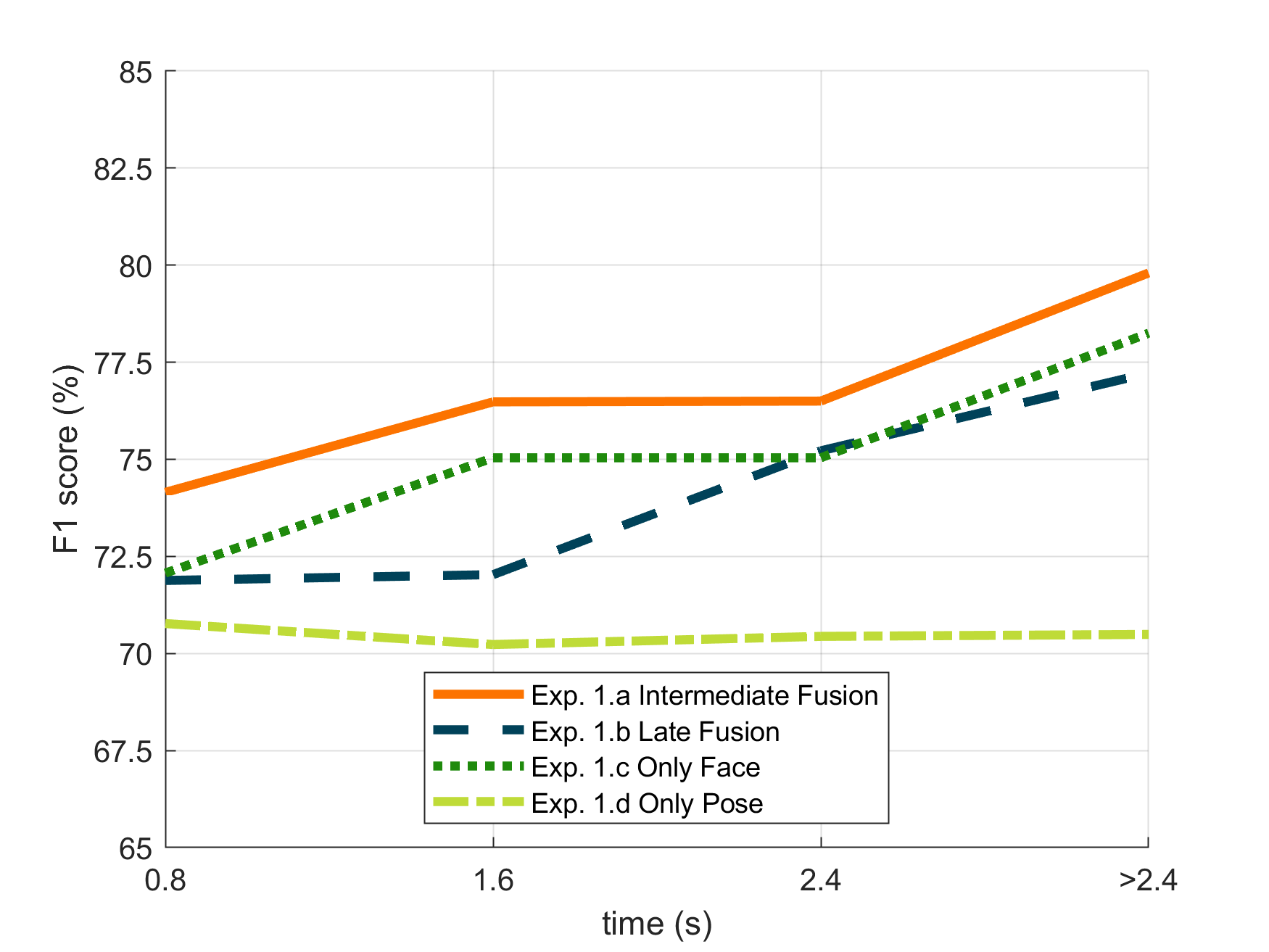}}
\caption{\textbf{Plot reporting performance of Addressee Estimation model as a function of the duration of the utterance.} Results of the 10-fold cross-validation experiments (Exp. 1.a-b-c-d) are provided in terms of mean of F1-score. Performance are computed at 0.8s, 1.6s, 2.4s, and for utterances lasting 2.4s or more. On the y-axis the performance score is expressed in \%.}
\label{fig:incremental_results}
\end{figure}
\section{Discussion} \label{ape_discussion}


In this work, Addressee Estimation has been conceived as the robot's ability to understand an utterance's addressee by interpreting the speaker's bodily cues. We tackle this problem as a three-class classification of the addressee's position with respect to the speaker. 

\subsection{The impact of the two visual input modalities on the model performance}
To solve this task, we designed a deep-learning model with convolutional layers and LSTM cells and we trained it with sequences of face images and body pose vectors of the speaker. As Skantze \cite{Skantze2020} reported, information from the speaker’s head pose, as a proxy of visual focus of attention, is highly relevant for humans when estimating others’ addressee and beneficial when implementing automatic Addressee Estimation models. Results from the testing phase of our Addressee Estimation model corroborated this perspective as the model could predict the position of the addressee. 

Interestingly, the model trained with only body pose information proved to be equally effective. It is true that also the body pose vectors contained information about the head direction, but this was gathered only by 5 key points: 1 for the nose, 2 for the eyes, and 2 for the ears. Compared to the face image, the body pose presented evidence of the speaker's whole-body direction. 

Though the difference was not substantial, the models trained with both visual modalities (face and body pose) performed better than the single-modality ones. This outcome was expected, as well as the fact that the intermediate-fusion approach resulted to be more effective than the late-fusion, as literature on the topic suggested \cite{Ramachandram2018, Stahlschmidt2022}.

The beneficial effect of combining the two visual input modalities may be explained by analysing the performance of each class more thoroughly. Although the overall performance of the two single-feature models was nearly identical, relevant differences appear when considering each class separately. As it appears in Figure \ref{fig:class_performances}, the only-face model predicts with higher F1-score the ``LEFT'' and ``RIGHT'' classes with respect to the ``ROBOT'' with a gap greater than 15\%. Different is the case for the only-pose model, whose performance is more stable along the three classes. What impacts more this result is the high recall for the ``ROBOT'', meaning that the model recognises the ``ROBOT'' more easily from body poses than face images. 

This situation seems to be reflected in the performance of the intermediate-fusion model, which combines a high performance for the ``LEFT'' and ``RIGHT'' classes with results more balanced for the ``ROBOT''. This indicates that beyond a general increase in performance given by the help of two channels instead of one, relevant features for ``LEFT'' and ``RIGHT'' are provided by the face modality, whereas for ``ROBOT'' by the body pose of the speaker. Interestingly, this pattern is not shared with the late-fusion model, in which the same gap between ``LEFT''/``RIGHT'' and ``ROBOT'' is even more evident than in the only-face model. Accordingly, this difference between the intermediate and the late-fusion approach suggests that fully-connected layers are not enough to optimally balance the two modalities.

In Exp. 2, we trained the model using both visual modalities combined with an intermediate fusion approach to solve a binary classification task and to compare our methods with the state-of-the-art model on the Vernissage Corpus \cite{Sheikhi2013}. Results of the overall-F1 score demonstrated that our approach slightly outperformed previous model. In addition, while in \cite{Sheikhi2013} the prediction required contextual information such as the possible targets of attention in the environment and could be obtained only at the end of the utterance, our model relies only on visual information automatically extracted from the robot's sensors and reliable results can be obtained just 0.8s after the start of the utterance.

\subsection{The longer you talk, the better I estimate}
Our Addressee Estimation model was developed so as to have predictions independent from the utterance length and available less than 1s after the utterance start. The method chosen to solve this task was to focus on sequences of data lasting 0.8s. For each utterance, this allowed providing a first prediction about the addressee at 0.8s, as well as other predictions every 0.8s after that, so that the final utterance prediction was incrementally weighted on all the predictions of sequences of that utterance. Longer utterances are formed by a higher number of sequences, hence more ample evidence for a correct estimate. This is what appears from results in Figure \ref{fig:incremental_results}. The longer the speaker talks, the better the estimate of the addressee. 

The intermediate-fusion, late-fusion, and only-face models share this pattern. Conversely, this is not the case for the only-pose model. It appears, therefore, that the two multimodal models inherit this characteristic from the only-face model. Moreover, one may speculate that the reason underlying this different behavior is that humans often turn their heads while speaking, in particular, if they are referring to some objects in the environment, as is the case in triadic interactions. The situation might be different for body poses that, although including information about the face pose, are more stable if considering the whole body, at least in the scenario of the Vernissage Corpus. 


\subsection{Addressee Estimation: a skill for social robots}
The present model is planned to be implemented on social robots. With this aim, we followed three general principles for its design: (i) keep the focus on speaker's bodily behavior, crucial to develop human-aware robots; (ii) integrate the temporal dimension of the task; (iii) produce an ecologically valid model, suitable for ecological scenarios.

Visual non-verbal information from the speakers' body (face and body pose) were used to achieve the prediction of their addressee. Our work highlights how important non-verbal behavior is to correctly interpret the meaning and the dynamics underlying verbal communication and advocates for making further use of this component for developing robots as conversational agents. Non-verbal behavior offers profound insights into other agents’ intentions. Targeting that is a valid solution to improve robot's skills of human awareness, crucial to enhance natural and effective HRI.

The idea of temporality influenced the design of the classification task and of the neural network. Firstly, temporality was conceived within each utterance. Utterances were not considered as a whole but partitioned in multiple time intervals of 0.8s, each of them generating a prediction. In this way, predictions about the addressee are available just after 0.8s the start of the utterance: a feature that reveals to be essential in case of real-time HRI. Secondarily, temporality was conceived within each sequence, i.e., within each time interval of 0.8s. The inputs of the neural network were not snapshots but temporal sequences connecting 10 frames. In this way, the final estimation does not only rely on instants but also on information related to the temporal dimension of the sequence, fundamental to interpret human behavior.

\begin{figure}[t!] 
\centerline{\includegraphics[width=1\columnwidth]{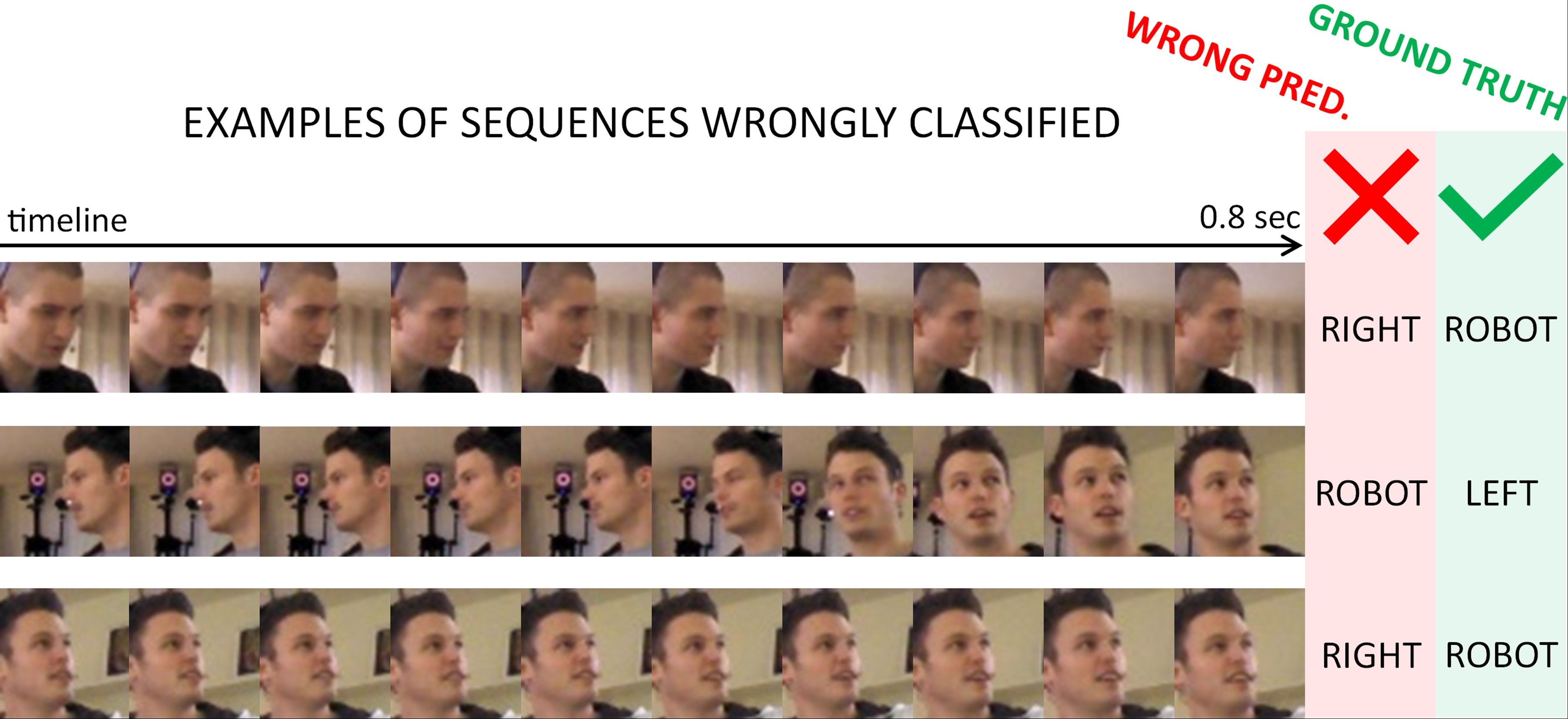}}
\caption{\textbf{Examples of sequences wrongly predicted.} The face images of four sequences are exhibited reporting the wrong prediction given by the intermediate-fusion model (Exp. 1.a) and the ground truth (correct addressee).}
\label{fig:errors}
\end{figure}

We chose the Vernissage Dataset to make the model suitable for ecological HRI scenarios. Since our Addressee Estimation model is designed and trained to be implemented on social robots, it was important to rely on data acquired directly through the robot's sensors and from its first-person perspective for its training. Moreover, using dataset recorded in embodied interaction with a physical robot was important because people behave differently in front of a physical robot \cite{Li2015}. In this respect, the Vernissage Corpus was optimal because it was also designed on triadic interactions, that is with a large part of the speech related to objects in the environments (in this case, the pictures on the wall). Instead of having a fixed conversational scenario (e.g., everybody sat at a table directly looking at their addressee), the Vernissage Corpus also includes situations more difficult to predict, but more common in ecological scenarios. Figure \ref{fig:errors} displays some sequences wrongly predicted by our model in Exp 1.a and shows that the speaker's head direction was not always predictive of the addressee’s position. On the contrary, in certain cases, it even caused errors. Triadic interactions are one of the causes of such errors because participants could talk to Nao while looking at a picture on the right or, vice versa, talking to their companion on the left while looking at a picture collocated over the robot. This demonstrates that approaches relying only on the speaker's visual focus of attention may have difficulties when it comes to provide predictions in `in-the-wild' scenarios. To overcome this difficulty, our approach combined information about the face and the whole body pose and balanced the predictions over time. In this way, our approach could achieve reliable results in situations more representative of `in-the-wild' interactions and outperform state-of-the-art models.

\subsection{Limitations and Future Work}
Our current Addressee Estimation model presents two major limitations. The model does not envisage the presence of a large number of people in the environment. A three-class classification of the position of the addressee would not be enough in case of crowded places such as airports, hotel reception areas, malls. However, this work is a step in advance with respect to most of the previous implementations of Addressee Estimation in HRI, which provide a binary output or require a fixed conversational scenario. Morover, tackling Addressee Estimation as prediction of the addressee's position may bring an additional benefit to the robot in terms of augmented perception of the external world. Since our model rely only on the speaker and no information about other agents is required, the information about the addressee's position may improve the robot’s awareness and localization of other agents not yet detected in the environment.

With regards to the second limitation, at the moment, the model has been designed only for visual information. No auditory or contextual information is used. By design, the model was conceived not to rely on contextual information, so that a first prediction of the addressee may be provided with the only use of visual information, without any knowledge about the number of agents in the environment, previous addressees, previous speakers, topic of the dialogue, hot-words, etc. Future works may add such contextual features to improve the performance of the model. Future implementations could also envisage the use of auditory cues, which comprise not only verbal information but also, for instance, prosody. Auditory data were already available in the Vernissage Corpus, but too noisy to extract reliable prosodic information.

\section{Conclusion}
In this work, we tackled the problem of Addressee Estimation, understanding to whom a speaker is addressing an utterance, by developing a deep learning hybrid model (CNN+LSTM) taking as input two visual information: the speaker's face and body pose. Our approach aims to develop a model for Addressee Estimation for real-time and ecological HRI. Our model achieves reliable performances even before the conclusion of the utterance and without any use of contextual information.
When it comes to develop social skills for robots, deep learning architectures prove to be a suitable solution but need to be designed combining the richness of information provided by human bodily behaviour, integrating the temporal dimension of the task, and considering the ecological validity of the model. In this direction, the present work represents a good stepping stone to be able to equip robots with Addressee Estimation skills and to enhance natural and effective HRI.




\bibliographystyle{IEEEtran}
\bibliography{IJCNN_Addressee_arXiv}

\end{document}